\documentclass[10pt,twocolumn,letterpaper]{article}

\usepackage{wacv}
\usepackage{times}
\usepackage{epsfig}
\usepackage{graphicx}
\usepackage{amsmath}
\usepackage{amssymb}

\def\wacvPaperID{311} 

\wacvfinalcopy 

\ifwacvfinal
\def\assignedStartPage{1} 
\fi

\ifwacvfinal
\usepackage[breaklinks=true,bookmarks=false]{hyperref}
\else
\usepackage[pagebackref=true,breaklinks=true,colorlinks,bookmarks=false]{hyperref}
\fi

\ifwacvfinal
\else
\fi

\begin{document}

\title{On the generalization of learning-based 3D reconstruction}

\author{Miguel Angel Bautista\\
Apple Inc.\\
\and
Walter Talbott \\
Apple Inc.\\
\and
Shuangfei Zhai \\
Apple Inc.\\
\and
Nitish Srivastava \\
Apple Inc.\\
\and
Joshua M. Susskind \\
Apple Inc.\\
{\tt\small \{mbautistamartin,wtalbott,szhai,nitish\_srivastava,jsusskind\}@apple.com}
}

\maketitle

\begin{abstract}
State-of-the-art learning-based monocular 3D reconstruction methods learn priors over object categories on the training set, and as a result struggle to achieve reasonable generalization to object categories unseen during training. In this paper we study the inductive biases encoded in the model architecture that impact the generalization of learning-based 3D reconstruction methods. We find that 3 inductive biases impact performance: the spatial extent of the encoder, the use of the underlying geometry of the scene to describe point features, and the mechanism to aggregate information from multiple views. Additionally, we propose mechanisms to enforce those inductive biases: a point representation that is aware of camera position, and a variance cost to aggregate information across views. Our model achieves state-of-the-art results on the standard ShapeNet 3D reconstruction benchmark in various settings.
\end{abstract}

\section{Introduction}

\begin{figure}[!t]
\centering
\includegraphics[width=0.48\textwidth]{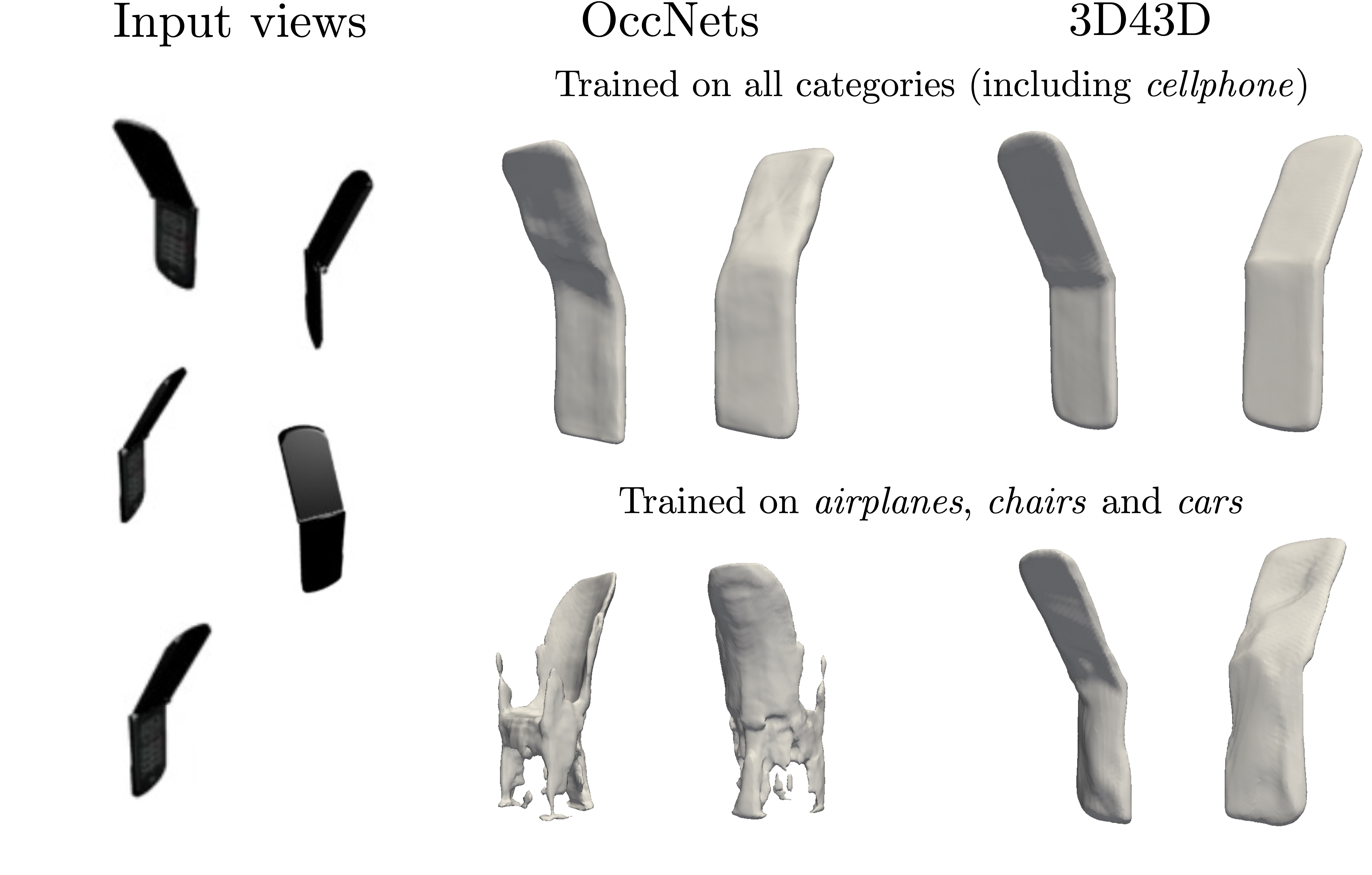}
\caption{An example of reconstructing object categories unseen during training.
State-of-the-art methods for learning-based reconstruction like OccNets \cite{occnet} fail to generalize to categories unseen during training, mapping objects to their closest category in the training set (\eg a chair). 3D43D improves generalization by using 3 inductive biases in the network design.}
\label{fig:intro}
\end{figure}

Reconstructing the 3D shape of an object from monocular input views is a fundamental problem in computer vision.  When the number of input views is small, reconstruction methods rely on priors over object shapes.  Learning-based algorithms encode such priors from data. Recently proposed approaches \cite{atlasnet,disn,occnet} have achieved success in the single/multi view, seen category case when generalizing to novel objects within the seen categories. However, these approaches have difficulty generalizing to object categories not seen during training (cf. Fig. \ref{fig:intro}).  

We present progress learning priors that generalize to unseen categories by incorporating a geometry-aware spatial feature map.  Within this paradigm, we propose a point representation aware of camera position, and a variance cost to aggregate information across views.

\begin{figure}[!t]
    \centering
    \includegraphics[width=0.48\textwidth]{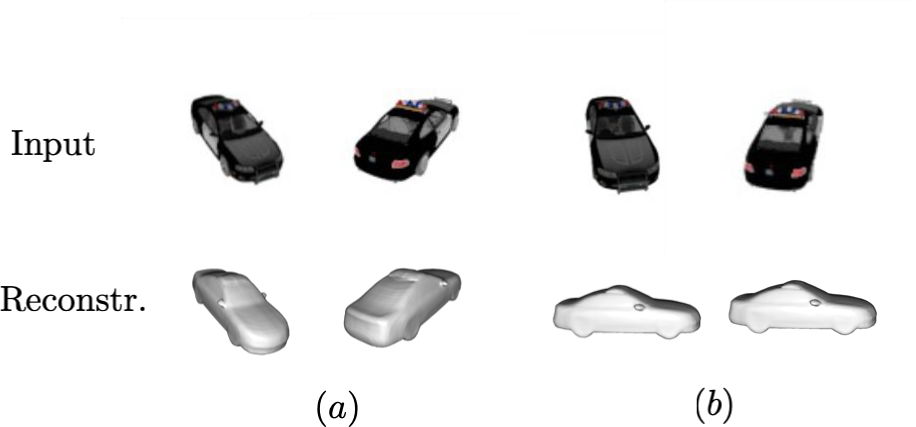}
    \caption{(a) View-centric coordinate system, where ground truth 3D objects are aligned to their respective input views. (b) Object-centric coordinate system, where all input views share the same ground-truth canonical 3D object orientation.}
    \label{fig:coords}
\end{figure}

A typical learning-based approach will take a single 2D view of an object as input, and a model to generate a 3D reconstruction. What should happen to the 3D ground truth as the viewpoint of the 2D input changes?  An \textit{object-centric} coordinate system would keep the ground truth fixed to a canonical coordinate system, regardless of the viewpoint of the 2D input view.  In contrast, a \textit{view-centric} coordinate system would rotate the ground truth coordinate frame in conjunction with the input view. An example of the two different coordinate systems~\footnote{In the graphics community the object-centric coordinate system is often referred as \textit{world coordinates} and the view-centric coordinate system as \textit{camera coordinates}.} is shown in Fig. \ref{fig:coords}. Object-centric coordinate systems align shapes of the same category to an arbitrary, shared coordinate system. This introduces stable spatial relationships during training (\textit{e.g.}, wheels of different car shapes generally occupy the same absolute area of $\mathbb{R}^3$). This makes the reconstruction task easier to learn, but these relationships are not necessarily shared across categories.  Similar to \cite{pixelsvoxelsviews,singleviewlearn}, we show empirically that adopting a view-centric coordinate system improves generalization to unseen categories.

Another critical factor for achieving good generalization to unseen categories is the capacity of a model to encode geometrical features when processing the input view. Similar to \cite{disn}, our model uses feature maps with spatial extent, rather than pooling across spatial locations to obtain a global representation. In \cite{disn} the motivation for using spatial feature maps is to preserve fine grained details of geometry (\eg to better model categories in the training set). In contrast, in this paper we analyze generalization to unseen categories and how different encoding designs impact generalization capability. We argue that using a globally pooled representation encourages the model to perform reconstruction at the object level (since the pooling operation is invariant to spatial arrangements of the feature map), which makes it difficult to generalize to objects from unseen categories. By keeping the spatial extent of features, on the other hand, we process and represent an object at the part level. Critically, in contrast to \cite{disn}, we model the scene geometry across different views by explicitly embedding information about camera poses in the spatial feature maps. We show empirically that using these geometry-aware spatial feature maps increases generalization performance.

Finally, we use multi-view aggregation to improve generalization performance.  Traditional approaches to 3D reconstruction, such as multi-view stereo (MVS) \cite{mvs} or structure-from-motion (SfM) \cite{sfm}, exploit the geometry of multiple views via cost volumes instead of priors learned from data.  These approaches fail in single-view cases.  Single-view reconstruction models, though, must rely entirely on priors for occluded regions.  We propose a model that combines learned priors with the complimentary information gained from multiple views. We aggregate information from multiple views by taking inspiration from cost volumes used in MVS and compute a variance cost across views. By refining its single-view estimates with additional views, our model shows improved generalization performance.


Individually, these factors are important as backed by literature and our empirical results and addressing them leads to compounding effects on generalization. The view-centric coordinate system has been shown to improve generalization \cite{pixelsvoxelsviews,singleviewlearn}. However, the need to aggregate information from multiple views is also paramount to reconstruct categories not seen during training time, since the prior learned over the training categories is not trustworthy in this unseen category case. In order to maximally benefit from aggregating information from multiple views, we require features that encode information about parts of objects rather than encoding the object as a whole entity without preserving spatial information. Otherwise, aggregation can be counterproductive by reinforcing the wrong object prior. We show empirically that by compounding these three factors, 3D43D outperforms state-of-the-art 3D reconstruction algorithms when tested on both categories seen and unseen during training. Contrary to suggestions from previous work \cite{unseenclasses,singleviewlearn}, we achieve these gains in generalization without a drop in performance when testing on categories seen during training. 

\begin{table*}[!t]
\scriptsize
    \centering
    \begin{tabular}{|c|c|c|c|c|c|c|c|c|}
    \hline
& MV Consist. \cite{tulsiani}  & Diff. PCs.  \cite{diffpcs}   & L-MVS \cite{lmvs} & PiFU \cite{pifu} & DISN \cite{disn} & OccNet \cite{occnet}  & 3D43D \\ \hline
    \textbf{Geometry} &  Voxel & Point cloud & Voxel/Depth & Func. & Func. & Func. & Func.\\ \hline
    \textbf{Coordinate Sys.} & Viewer & Viewer & Object & Object & Object & Object & Viewer \\ \hline
    \textbf{Features}  & Global & Global & Spatial+Global & Spatial+Global & Spatial+Global & Global & Spatial+Geometric \\ \hline
    \textbf{Multi-view} & No & No & Yes & Yes & Yes & No & Yes\\ \hline
    \textbf{Generalization} & No & No & Yes & No & No & No & Yes\\ \hline
    \end{tabular}
    \caption{Summary of design choices of different approaches. We describe each method in terms of their choice of: geometry representation (\textbf{Geometry}), coordinate system (\textbf{Coordinate Sys.}), feature representation (\textbf{Features}), capacity to use multiple views (\textbf{Multi-view}) and if they analyze generalization to unseen categories (\textbf{Generalization}).}
    \label{tab:method_summarization}
\end{table*}

\section{Related Work}

\textbf{Object Coordinate System.} 
Careful and extensive experimental analysis in \cite{pixelsvoxelsviews,singleviewlearn} has revealed that the object shape priors learned by most 3D reconstruction approaches act in a categorization regime rather than in the expected 3D dense reconstruction regime. In other words, reconstructing the 3D object in these models happens by categorizing the input into a blend of the known 3D objects in the training set. This phenomenon has been mainly explained by the choice of the object-centric coordinate system. 

Results by \cite{pixelsvoxelsviews,singleviewlearn} showed that object-centric coordinate systems perform better for reconstructing viewpoints and categories seen during training, at the cost of significantly hampering the capability to generalize to categories or viewpoints unseen at training time. The converse result was also observed for view-centric coordinate systems, which generalized better to unseen objects and viewpoints at the cost of degraded reconstruction performance for known categories/viewpoints.

\textbf{Feature Representation.} Single-view 3D reconstruction approaches have recently obtained impressive results \cite{occnet,pifu,disn,tulsiani,atlasnet, softraster,birds} despite the ill-posed nature of the problem. In order for these approaches to perform single-view reconstruction successfully, priors are required to resolve ambiguities. Not surprisingly, recent works show that using local features that retain spatial information \cite{pifu,disn} improve reconstruction accuracy. However, none of these approaches analyze their performance on object categories unseen during training time. A recent exception is \cite{unseenclasses} where the authors propose a \textit{non} fully differentiable approach for single view reconstruction that relies on depth in-painting of spherical maps \cite{sphericalmaps}. \cite{unseenclasses} also differs from 3D43D because 3D voxel grids are used as ground-truth, and extra supervision at the level of depth maps is available at training time.

\textbf{View Aggregation.} Multi-view 3D reconstruction has been traditionally addressed by stereopsis approaches like MVS \cite{mvs,mvs2} or SfM \cite{sfm}. Modern learning-based approaches to MVS \cite{mvsnet,mvs2} have incorporated powerful convolutional networks to the MVS pipeline.  These networks focus on visible regions and do not make inferences about the geometry of occluded object parts. 

Another interesting trend has been to exploit the multi-view consistency inductive bias from MVS to learn 3D shape and pose from pairs of images \cite{tulsiani,diffpcs,keypointnet}. However, these approaches either predict a very sparse set of keypoints \cite{keypointnet}, a sparse point cloud \cite{diffpcs}, or a voxel grid \cite{tulsiani}, limiting the approaches to fixed resolution representations. 

Conceptually close to our approach is \cite{lmvs}. The authors propose differentiable proxies to the operations in the standard MVS pipeline, allowing end-to-end optimization.  Although \cite{lmvs} addresses multi-view aggregation, there are critical design choices in other aspects of the method that limit the performance. First, the geometry representation produced is a voxel grid, making the estimation of high resolution geometry unfeasible. Second, the cost-volume optimization happens via a large 3D auto-encoder which has a non trivial geometric interpretation. Third, view aggregation is performed in a recurrent fashion, making the model sensitive to permutation of the views. 

Properly extending the previously discussed single-view works \cite{occnet,pifu, disn,tulsiani,atlasnet,softraster,birds} to the multi-view case is not trivial, although simple extensions to aggregate multiple views are briefly outlined in \cite{pifu, disn}.  Inspired by cost volume computation used in MVS \cite{mvsnet} we aggregate information from different views by computing a variance cost (Sect. \ref{sec:muli-view}.)

\textbf{Geometry Representation.} The choice of representation scheme for 3D objects has been at the core of 3D reconstruction research from the early beginning. Voxels \cite{tulsiani,lmvs} have been used as a natural extension of 2D image representation, showing great results in low resolution regimes. However, memory and computation requirements to scale voxel representations to higher resolution prevent them from being widely used. Circumventing this problem, point clouds are a more frequently used representation \cite{diffpcs,keypointnet}. Point clouds deal with the computational cost problem of voxels by sparsifying the representation and eliminating the neighbouring structure information. Meshes \cite{birds,meshrenderer,softraster} add the neighboring structure back into point cloud representations. However, to make mesh estimation efficient, neighbouring structure has to be predefined (usually in the form of connectivity of a convex polyhedron with a variable number of faces) and only deformations of that structure can be modelled. Finally, functional/implicit representations have recently gained interest \cite{occnet,disn,deepsdf,deeplevelsets,cvxnet,nasa}. This representation encodes geometry as the level set of a function that can be evaluated for any point in $\mathbb{R}^3$.  Such a function can generate geometry at arbitrary resolutions by evaluating as many points as desired. As a summary, Tab. \ref{tab:method_summarization} shows the contributions of the most relevant and related literature in comparison to 3D43D.

\section{Model}

We now describe our approach in terms of the choice of geometry representation, the use of geometry-aware feature maps, and the multi-view aggregation strategy. Our model is shown in Fig. \ref{fig:model}. 

\subsection{Functional Geometry Representation}

3D43D takes the form of a functional estimator of geometry \cite{occnet,disn,atlasnet,deepsdf}. Given a view of an object, our goal is to predict the object occupancy function indicating whether a given point $\mathbf{p} \in \mathbb{R}^{3}$ lies inside the mesh of the object. In order to do so, we learn a parametric scalar field $f_\theta: \mathbb{R}^3 \times \mathbf{V} \longrightarrow [0, 1]$ where $\mathbf{V} \in \mathbb{R}^{H \times W \times 3}$ is an monocular (RGB) view of the object. In the remainder of the text the parameter subscript $\theta$ is dropped for ease of notation. This scalar field $f$ is implemented by a fully connected deep neural network with residual blocks ~\footnote{Details of the implementation can be found in the supplementary material}.

\subsection{Encoding Geometry-aware Features}\label{sect:encoding}

Our goal is to learn a prior over occupancy functions that generalizes well to unseen categories, which we address by giving our model the capacity to reason about local object cues and scene geometry. In order to do so, we process input views with a convolutional U-Net\cite{unet} encoder with skip connections (refer to the supplementary material for implementation details). This results in a feature map $\mathbf{C} \in \mathbb{R}^{H \times W \times C}$ for a given RGB view. This is in contrast to the approach taken in \cite{occnet, atlasnet} where a view is represented by a global feature which pools across the spatial dimensions of the feature map. Our hypothesis is that preserving the spatial dimensions in the latent representation is crucial to retain local object cues which greatly improve generalization to unseen views and categories, as demonstrated in the experiments. To predict the occupancy value for a 3D point $\mathbf{p}$ in world coordinates we project this point into its location $(u, v)$ in $\mathbf{C}$ by using the extrinsic camera parameters $\mathbf{T}=[\mathbf{R}|\mathbf{t}] \in \mathbb{R}^{3 \times 4}$ and intrinsic parameters $\mathbf{K} \in \mathbb{R}^{3 \times 3}$ ~\footnote{We assume camera intrinsics to be constant.} (cf. Eq. \eqref{eq:intrinsic}), and sample the corresponding feature vector $\mathbf{c} \in \mathbb{R}^C$. We use bi-linear sampling as in \cite{spatialtransformer} to make the sampling operation differentiable. 

\begin{eqnarray}
\label{eq:intrinsic}
    (u', v', w') = \mathbf{K}\mathbf{T}\mathbf{p}, \;\; u = u'/w', \;\;
    v = v'/w'.
\end{eqnarray}

The feature vector $\mathbf{c}$ encodes localized appearance information but lacks any notion of the structural geometry of the underlying scene. This causes ambiguities in the representation since, for example, all points in a ray shot from the camera origin get projected to the same location in the image plane. Thus, the sampled feature vector $\mathbf{c}$ cannot uniquely characterize the points in the ray (\eg, to predict occupancy). Recent works \cite{pifu,disn} mitigate this issue by augmenting $\mathbf{c}$ to explicitly encode coordinates of 3D points $\mathbf{p}$. This is often done by concatenating $\mathbf{p}$ (or a latent representation of $\mathbf{p}$ \cite{disn}) and $\mathbf{c}$, and further processing it via additional fully connected blocks.

However, $\mathbf{p}$ (or its representation) is sensitive to the choice of coordinate system \cite{pixelsvoxelsviews}. Recent approaches \cite{occnet,disn} use a canonical object-centric coordinate system for each object category, which has been shown to generalize poorly to categories not seen during training \cite{singleviewlearn}. On the other hand, expressing $\mathbf{p}$ in a view-centric coordinate system\footnote{Also known as camera coordinate system} improves generalization to unseen categories \cite{pixelsvoxelsviews,singleviewlearn,unseenclasses}. Note that if $\mathbf{p}$ is expressed in the view-centric coordinate system the characterization of the scene is incomplete since it lacks information about the points where the rays passing through $\mathbf{p}$ originated in the image capturing process (\eg the representation is not aware of the origin of the view-centric coordinate system \wrt the scene).

\begin{figure*}[!t]
\centering
\includegraphics[width=\textwidth]{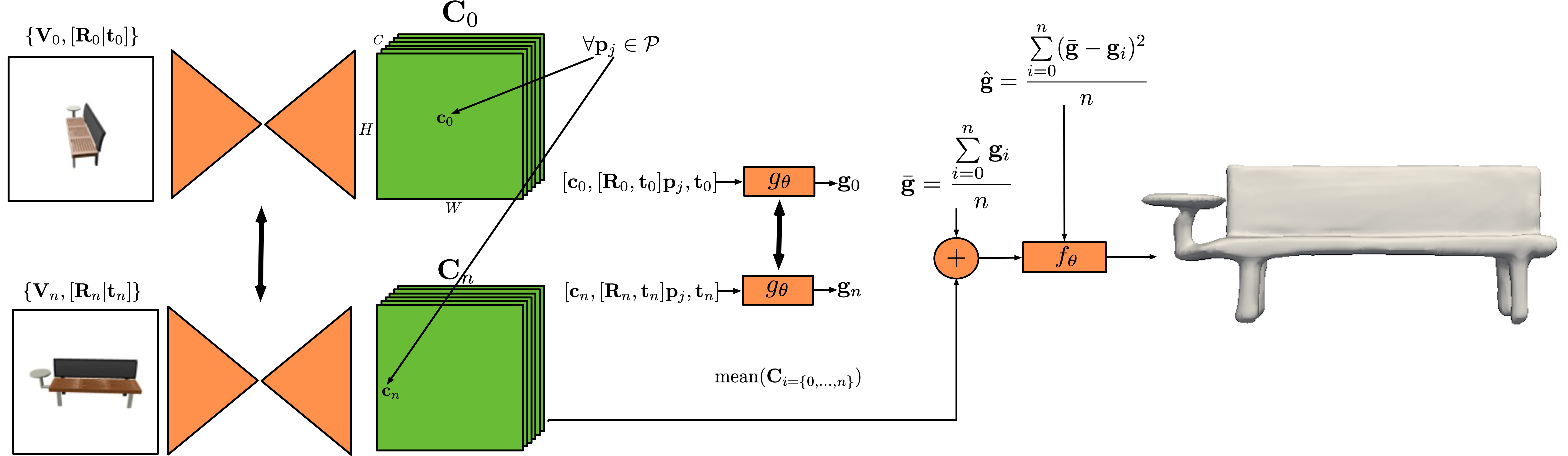}
\caption{Overview of our model. Input views $\textbf{V}_i$ are processed by our UNet encoder producing feature maps $\textbf{C}_i$ that are sampled at spatial locations corresponding to a 3D point $\textbf{p}_j$. Those features are then concatenated with the point $\textbf{p}_j$ and the location of the camera origin of the corresponding input view and process through an MLP $g_\theta$ that produces geometry-aware point representations (one for each view). Those representations are used to compute a mean and variance cost across views that is used by another MLP $f_\theta$ to predict occupancy.}
\label{fig:model}
\end{figure*}

To tackle this issue we represent $\mathbf{p}$ using the camera coordinate system (denoted as $\mathbf{p}'$), and give the representation access to the origin of the camera coordinate system $\mathbf{t} \in \mathbb{R}^3$ with respect to the world (\eg the camera position with respect to the world coordinate system). Therefore, after sampling $\mathbf{c}$ we concatenate it with $\mathbf{p}'$ and $\mathbf{t}$, and process it with an MLP $g_\theta: \mathbb{R}^n \rightarrow \mathbb{R}^n $ with residual blocks, resulting in feature representation $\mathbf{g}$ that is aware of the scene geometry $\mathbf{g} = g_\theta([\mathbf{c}, \mathbf{p}',\mathbf{t}])$ . This feature representation $\mathbf{g}$ is then input to the occupancy field $f$.  Note that this does not require additional camera information compared to \cite{disn,pifu} since the camera position is already used to project $\mathbf{p}$ into the image plane to sample the feature map. In our model we explicitly condition the representation using the camera position instead of only implicitly using the camera position to sample feature maps. Fig. \ref{fig:model} shows our model.

\subsection{Multi-View Aggregation} 
\label{sec:muli-view}

We now turn to the task of aggregating information from multiple views to estimate occupancy. Traditionally, view aggregation approaches for geometry estimation require the explicit computation of a 3D cost volume which is either refined using graph optimization techniques in traditional MVS approaches \cite{mvs} or processed by 3D convolutional auto-encoders with a large number of parameters in learned models \cite{lmvs,mvsnet,mvs2}. Here we do not explicitly construct a cost volume for the whole scene, instead, we compute point-wise estimates of that cost volume. One key observation is that our model is able to estimate geometry for parts of the object that are occluded in the input views, as opposed to MVS approaches that only predict geometry for visible parts of a scene (\eg depth). As a result our approach integrates \textit{reconstruction of visible geometry} and \textit{generation of unseen geometry} under the same framework.

Our task is to predict the ground truth occupancy values $\mathbf{o} \in \{0, 1\}^p$ of points $\mathcal{P}=\{\mathbf{p}_j\}_{j=0}^p$, given a set of posed RGB views $\{\mathbf{V}_i, \mathbf{T}_i\}_{i=0}^n$. In order to do so, we independently compute geometry aware representations $\mathbf{g}_{i}$ across views for each point $\mathbf{p}_j$ as show in Sect. \ref{sect:encoding}. In order for our model to deal with a variable number of views a pooling operation over $\mathbf{g}_{i}$ is required. Modern approaches to estimate the complete geometry of a scene (visible and occluded) from multiple views rely on element-wise pooling operators like mean or max \cite{disn,pifu}. These element-wise operators can be suitable for categories seen at training time. However, in order to better generalize to unseen categories it is oftentimes beneficial to rely on the alignment cost between features as done in purely geometric (\eg non learning-based) approaches \cite{mvs}. Inspired by traditional geometric approaches we propose to use an approximation to the alignment cost between local features $\{\mathbf{g}_{i}\}_{i=0}^n$. We approximate the alignment cost on the set of local features $\{\mathbf{g}_{i}\}_{i=0}^n$ by computing the variance $\hat{\mathbf{g}}$ as follows,

\begin{equation}
\label{eq:variance}
    \hat{\mathbf{g}} = \mathcal{V}(\{\mathbf{g}_i \}_{i=0}^n)= \frac{ \sum\limits_{i=0}^{n}\|\bar{\mathbf{g}} - \mathbf{g}_i\|_2}{n},
\end{equation}

where $\bar{\mathbf{g}}$ is the average of $\{\mathbf{g}_i\}_{i=0}^n$. A key design choice is that we do not use the variance $\hat{\mathbf{g}}$ as the sole input to our functional occupancy decoder $f$ since the variance will be zero everywhere and uninformative when only a single view is available. Instead, we add a conditional input branch to our decoder $f$, which takes as input $\bar{\mathbf{g}}$ conditioned on $\hat{\mathbf{g}}$. We also give the model access to a global object representation by introducing a residual path that performs additive conditioning on $\bar{\textbf{g}}$. We perform average pooling on feature maps $\textbf{C}_i$ both spatially and across views to obtain $\bar{\textbf{c}}$ that is added to $\bar{\textbf{g}}$. Conditioning in $f$ is implemented via conditional batch normalization\footnote{Implementation details in the supplementary material} layers \cite{cbn}. This formulation naturally handles the single view case, where $\hat{\mathbf{g}}=0$. Finally, our objective function is shown in Eq.~\ref{eq:loss}.

\begin{equation}
\label{eq:loss}
L(\{\mathbf{V}_i, \mathbf{T}_i\}_{i=0}^{n}, \mathcal{P}, \mathbf{o}) = - \sum\limits_{j=1}^{p} o_j \log f( \bar{\mathbf{g}} + \bar{\textbf{c}}, \hat{\mathbf{g}})
\end{equation}

\section{Experiments}

We present empirical results that show how 3D43D performs in two experimental setups: reconstruction of categories seen during training and generalization to categories unseen during training. In the first setup our goal is to show that 3D43D is competitive with state-of-the-art 3D reconstruction approaches. In the second setup we show that 3D43D generalizes better to categories unseen at training time. Finally, we conduct ablation experiments to show how the proposed contributions impact the reconstruction accuracy.

\subsection{Settings}

\textbf{Dataset}: For all of our experiments, we use the ShapeNet \cite{shapenet} subset of Choy et al. \cite{3dr2n2}, together with their renderings. For a fair comparison with different methods we use the same train/test splits and occupancy ground truth as \cite{occnet}, which provides an in depth comparison with several approaches~\footnote{Readers interested in the ground-truth generation process are referred to \cite{occnet,stutz2018learning}.}.

\textbf{Metrics}: We report the following metrics, following \cite{occnet}: volumetric IoU, Chamfer-L1 distance, and normal consistency score. In addition, as recently suggested by \cite{singleviewlearn} we report the F-score. Volumetric IoU is defined as the intersection over union of the volume of two meshes. An estimate of the volumetric IoU is computed by randomly sampling 100k points and determining whether points reside in the meshes \cite{occnet}. The Chamfer-$L_1$ is a relaxation of the symmetric Hausdorff distance measuring the average of reconstruction accuracy and completeness. The normal consistency score is the mean absolute dot product of the normals in one mesh and the normals at the corresponding nearest neighbors in the other mesh \cite{occnet}. Finally, the F-score can be understood as “the percentage of correctly reconstructed surface” \cite{singleviewlearn}.

\textbf{Implementation}: We resize our input images to $(224, 224)$ pixels. For our encoder, we choose a U-Net with a ResNet-50\cite{resnet} encoder, the final feature maps $\textbf{C}$ have $256$ channels and are of the same spatial size as the input. The function $g_\theta$ that computes geometric features is an MLP with $3$ ResNet blocks and that takes a vector of $262=256+3+3$ dimensions and outputs a $256$-dimensional representation $\textbf{g}$. Our occupancy function $f$ is an MLP with $5$ ResNet blocks where all layers have $256$ hidden units except the output layer. To train the occupancy function we sample $2048$ points with their respective occupancy value from a pool of $100$k points and use $4$ input views \footnote{This was due to memory limitations. Nonetheless, the method generalizes to an arbitrary number of input views during inference}. Details of different sampling strategies can be found in \cite{occnet}. We train our network with batches of $128$ samples and use Adam\cite{adam} with default Pytorch parameters as our optimizer. We use a learning rate of $10^{-4}$ and train our network for $2000$ epochs. To obtain meshes at inference time from the occupancy function we follow the process in \cite{occnet}.

\subsection{Categories seen during training}

In this section, we compare our method to various baselines on single-view and multi-view 3D reconstruction experiments. For the single-view setup we report our results on standard experiments for reconstructing unseen objects of categories seen during training (cf. Tab. \ref{tab:all_category}). We compare 3D43D with 3D-R2N2\cite{3dr2n2}, Pix2Mesh\cite{pix2mesh}, PSGN\cite{psgn}, AtlasNet\cite{atlasnet} and OccNets\cite{occnet}. Encouragingly, 3D43D performs on par with the state-of-the-art OccNets\cite{occnet} approach, which uses global features that are able to encode semantics of the objects in the training set. In addition, OccNets make use of an object-centric coordinate system, which aligns all shapes to the same canonical orientation, making the reconstruction problem simpler. Our results indicate that spatial feature maps are able to encode useful information for reconstruction despite being spatially localized. This result backs up our initial hypothesis and is critical to establish a good baseline performance.  From this baseline, we explore the performance of our model when generalizing to unseen categories in different scenarios (Sect. \ref{sect:unknown_categories}).

\begin{table*}[!t]
\centering
\scriptsize
\begin{tabular}{|l|cccc|cccccc|ccccc|}
\hline

{} & \multicolumn{4 }{|c|}{\textbf{IoU}} & \multicolumn{6}{|c|}{$\downarrow$\textbf{Chamfer-$L_1$}} & \multicolumn{5}{|c|}{\textbf{Normal Consistency}} \\ 

\textbf{Seen category} & \cite{3dr2n2} &        \cite{pix2mesh}  &            \cite{occnet}  & 3D43D&       \cite{3dr2n2} &   \cite{psgn} & \cite{pix2mesh} &        \cite{atlasnet} &            \cite{occnet}  & 3D43D &           \cite{3dr2n2} & \cite{pix2mesh} &        \cite{atlasnet} &           \cite{occnet}   & 3D43D\\ 
 \hline

airplane&   0.426   &   0.420   &   \textbf{0.591}  &   0.571   &   0.227   &   0.137   &   0.187   &   0.104  &   0.134   &  \textbf{ 0.096}   &   0.629   &   0.759   &   0.836   &   \textbf{0.845}  &   0.825   \\
bench   &   0.373   &   0.323   &   0.492   &   \textbf{0.502}  &   0.194   &   0.181   &   0.201   &   0.138  &   0.150   &   \textbf{0.112}   &   0.678   &   0.732   &   0.779   &   \textbf{0.814}  &   0.809\\
cabinet &   0.667   &   0.664   &   0.750   &   \textbf{0.761}  &   0.217   &   0.215   &   0.196   &   0.175   &   0.153   &   \textbf{0.119}  &   0.782   &   0.834   &   0.850   &   0.884   &   \textbf{0.886}\\
car     &   0.661   &   0.552   &   \textbf{0.746}  &   0.741   &   0.213   &   0.169   &   0.180   &   0.141   &   0.149   &   \textbf{0.122}  &   0.714   &   0.756   &   0.836   &   \textbf{0.852}  &   0.844\\
chair   &   0.439   &   0.396   &   0.530   &   \textbf{0.564}  &   0.270   &   0.247   &   0.265   &   0.209   &   0.206   &   \textbf{0.193}  &   0.663   &   0.746   &   0.791   &   0.829   &   \textbf{0.832}\\
display &   0.440   &   0.490   &   0.518   &   0.548   &   \textbf{0.605}  &   0.284   &   0.239   &   0.198   &   0.258   &   \textbf{0.166}  &   0.720   &   0.830   &   0.858  &   0.857   &   \textbf{0.883}\\
lamp    &   0.281   &   0.323   &   0.400   &   \textbf{0.453}  &   0.778   &   0.314   &   0.308   &   \textbf{0.305}  &   0.368   &   0.561   &   0.560   &   0.666   &   0.694   &   0.751   &   \textbf{0.766}\\
loudspeaker &   0.611   &   0.599   &   0.677   &   \textbf{0.729}  &   0.318   &   0.316   &   0.285   &   0.245  &   0.266   &   \textbf{0.229}   &   0.711   &   0.782   &   0.825   &   0.848   &   \textbf{0.868}\\
rifle   &   0.375   &   0.402   &   0.480   &   \textbf{0.529}  &   0.183   &   0.134   &   0.164   &   \textbf{0.115}  &   0.143   &   0.248   &   0.670   &   0.718   &   0.725   &   0.783   &   \textbf{0.798}\\
sofa    &   0.626   &   0.613   &   0.693   &   \textbf{0.718}  &   0.229   &   0.224   &   0.212   &   0.177   &   0.181   &   \textbf{0.125}  &   0.731   &   0.820   &   0.840   &   0.867   &   \textbf{0.875} \\
table   &   0.420   &   0.395   &   0.542   &   \textbf{0.574}  &   0.239   &   0.222   &   0.218   &   0.190   &   0.182   &   \textbf{0.146}   &   0.732   &   0.784   &   0.832   &   0.860   &   \textbf{0.864} \\
telephone   &   0.611   &   0.661   &   \textbf{0.746}  &   0.740   &   0.195   &   0.161   &   0.149   &   0.128   &   0.127   &   \textbf{0.107}  &   0.817   &   0.907   &   0.923   &   \textbf{0.939}   &   0.935 \\
vessel  &   0.482   &   0.397   &   0.547   &   \textbf{0.588}  &   0.238   &   0.188   &   0.212   &   \textbf{0.151}  &   0.201   &   0.175   &   0.629   &   0.699   &   0.756   &   0.797   &   \textbf{0.799}\\
\hline
mean    &   0.493   &   0.480   &   0.593   &   \textbf{0.621}  &   0.278   &   0.215   &   0.216   &   \textbf{0.175}  &   0.194   &   0.184   &   0.695   &   0.772   &   0.810   &   0.840   &   \textbf{0.845}\\ \hline
\end{tabular}
\caption{Performance of different approaches on the test set of categories seen during training, trained with single views. Our results show that 3D43D is comparable with state-of-the-art models trained on a object-centric coordinate system in the single view setting. Compared models are: 3D-R2N2\cite{3dr2n2}, Pix2Mesh\cite{pix2mesh}, PSGN\cite{psgn}, AtlasNet\cite{atlasnet} and OccNets\cite{occnet}.}
\label{tab:all_category}
\end{table*}

\begin{figure*}[!t]
\centering
\begin{tabular}{ccc}
\includegraphics[width=0.32\textwidth]{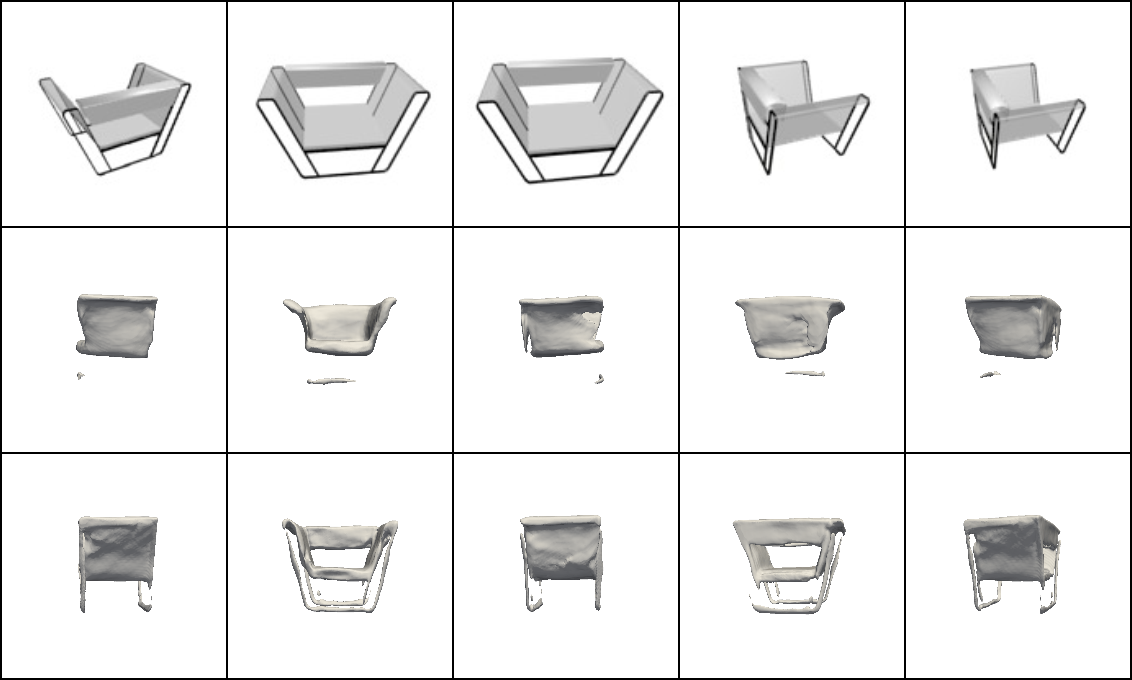} & \includegraphics[width=0.32\textwidth]{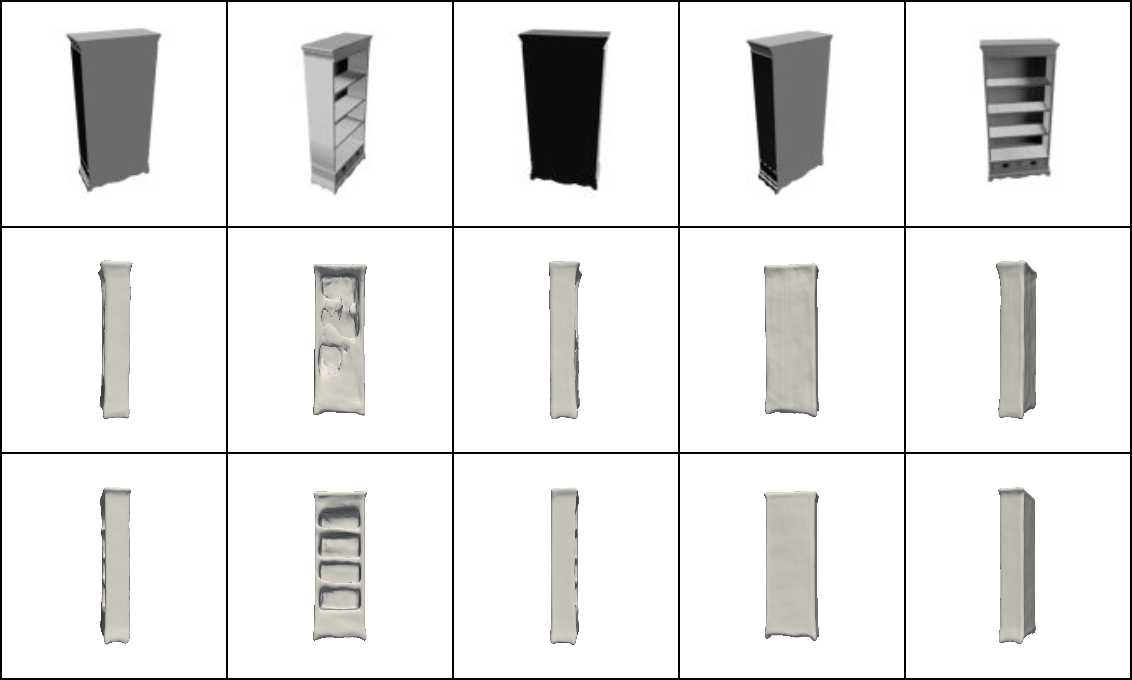} & \includegraphics[width=0.32\textwidth]{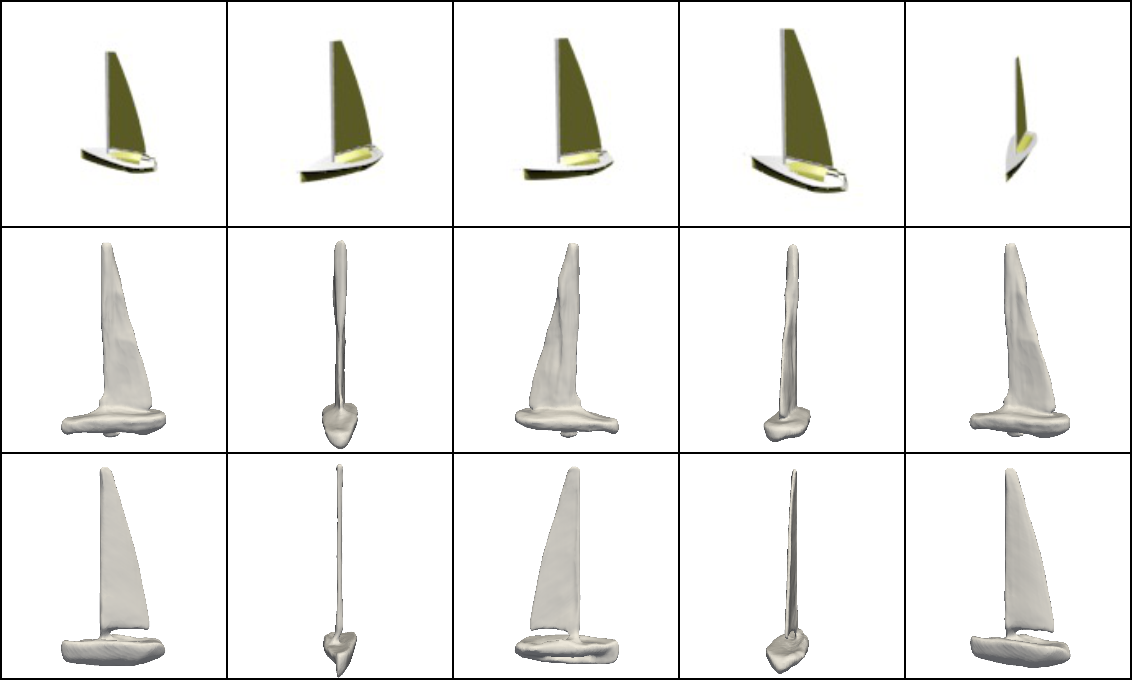} \\

\includegraphics[width=0.32\textwidth]{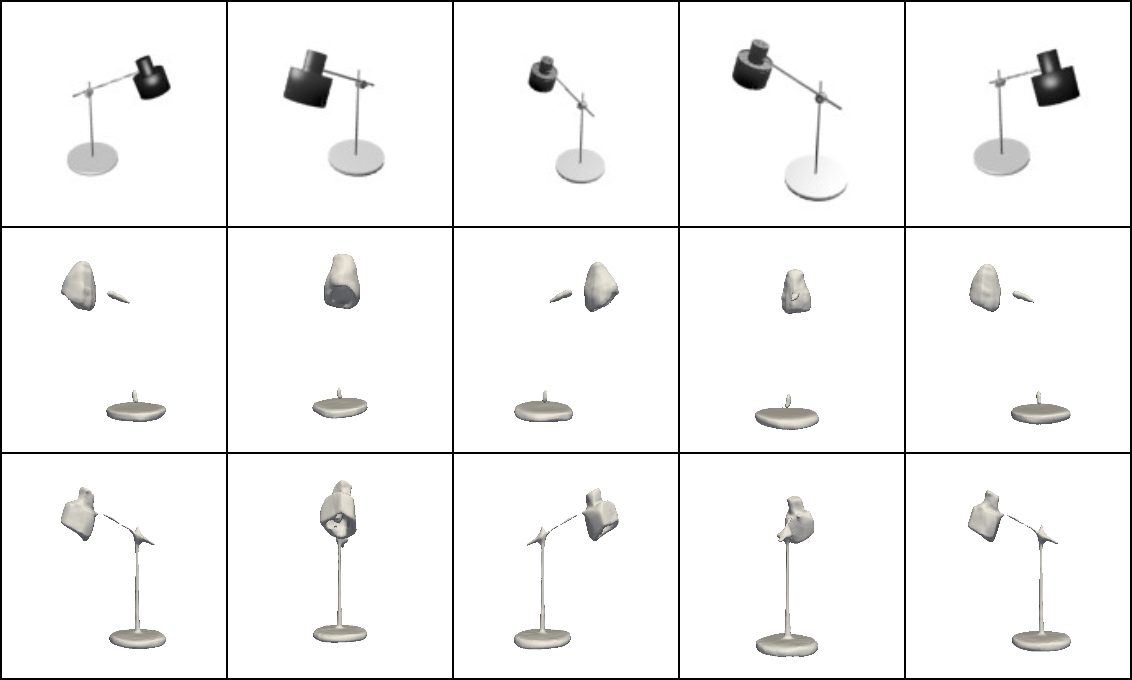} & \includegraphics[width=0.32\textwidth]{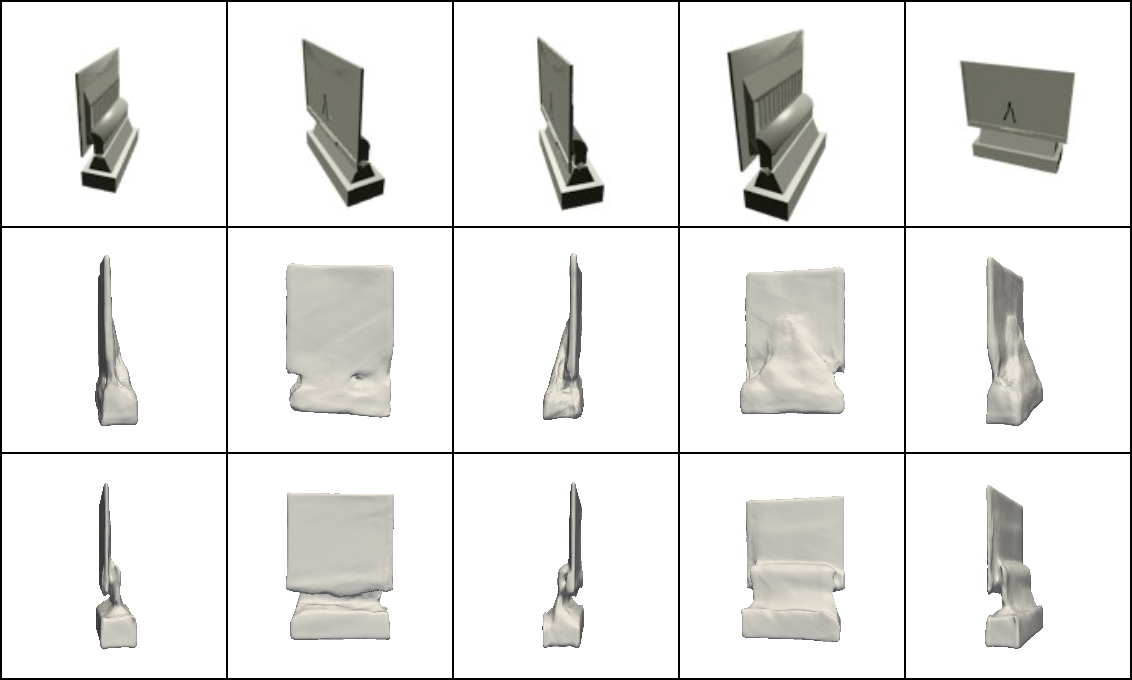} & \includegraphics[width=0.32\textwidth]{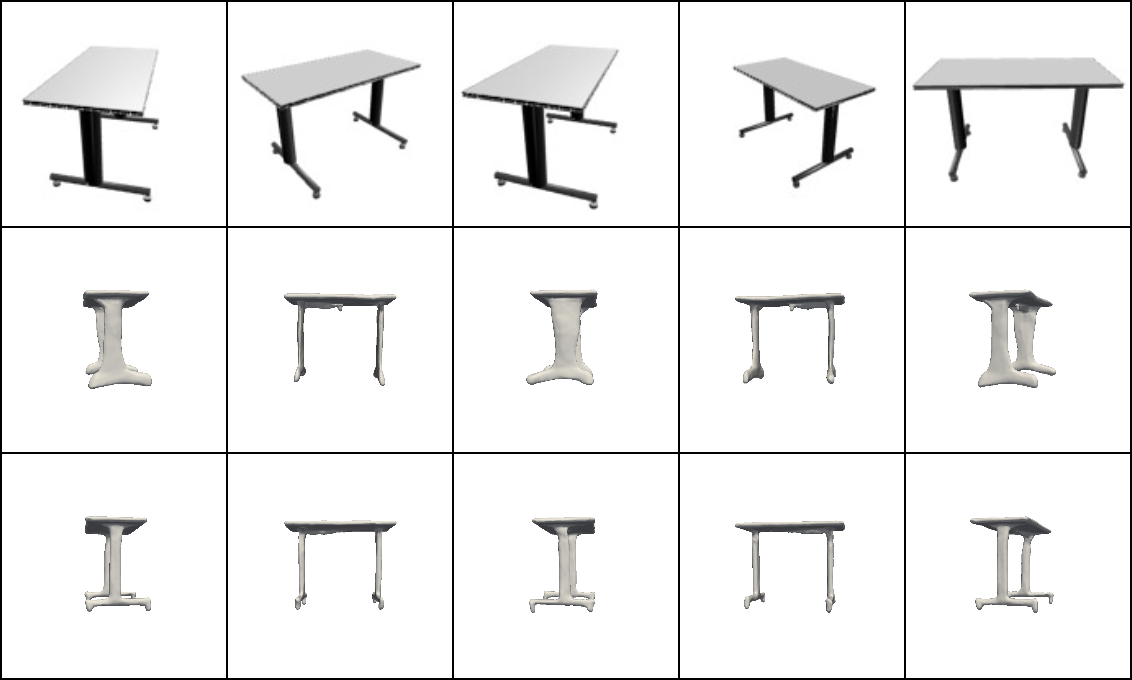} \\
\end{tabular}

\caption{Reconstructions from categories seen during training time using 5 input views. For each object: (\textbf{Top row}) Input views. (\textbf{Middle row}) OccNets \cite{occnet} prediction (orbit of 5 views of the predicted mesh). (\textbf{Bottom row}) 3D43D prediction (orbit of 5 views of the predicted mesh). We can qualitatively see that 3D43D produces better results than OccNets \cite{occnet} in terms of high-frequency geometry. Note that input views and reconstructions are not presented from the shown viewpoint.}
\label{fig:seen_categories}
\end{figure*}

In order to further validate our contribution, we provide results on multi-view reconstruction. We randomly sample 5 views of the object and compare our method with OccNets\cite{occnet} (the top performer for single-view reconstruction method). To provide a fair comparison, we extend the trained model provided by OccNets (the best runner up) to the multiple view case by average pooling their conditional features across views at inference time. Since our method uses spatial features that are aware of scene geometry, we expect our aggregation mechanism to obtain more accurate reconstruction. Results shown in Tab. \ref{tab:seen_categories_5_view} and qualitative results in Fig. \ref{fig:seen_categories} consistently agree with this observation. 

\begin{table*}[!t]
\scriptsize
\centering
\begin{tabular}{|l|cc|cc|cc|cc|}
\hline
& \multicolumn{2}{|c|}{\textbf{IoU}} & \multicolumn{2}{|c|}{$\downarrow$ \textbf{Chamfer-}$L_1$} & \multicolumn{2}{|c|}{\textbf{Normal Consistency}} & 
\multicolumn{2}{|c|}{\textbf{F-score}}\\ 
\textbf{Seen category }& \cite{occnet} & 3D43D & \cite{occnet} & 3D43D & \cite{occnet} & 3D43D & \cite{occnet} & 3D43D \\ \hline

    airplane &  0.600 &  \textbf{0.736} &         0.096 &         \textbf{0.021} &                0.853 &                \textbf{0.899} &     0.735 &     \textbf{0.841} \\
       bench &  0.547 &  \textbf{0.663} &         0.176 &         \textbf{0.027} &                0.834 &                \textbf{0.881} &     0.691 &     \textbf{0.789} \\
     cabinet &  0.770 &  \textbf{0.831} &         0.125 &         \textbf{0.073} &                0.893 &                \textbf{0.925} &     0.853 &     \textbf{0.898} \\
         car &  0.759 &  \textbf{0.797} &         0.109 &         \textbf{0.090} &                0.861 &                \textbf{0.873} &     0.852 &     \textbf{0.878} \\
       chair &  0.568 &  \textbf{0.716} &         0.187 &         \textbf{0.063} &                0.846 &                \textbf{0.911} &     0.704 &     \textbf{0.824} \\
     display &  0.593 &  \textbf{0.752} &         0.168 &         \textbf{0.089} &                0.884 &                \textbf{0.935} &     0.723 &     \textbf{0.851} \\
        lamp &  0.415 &  \textbf{0.625} &         1.083 &         \textbf{0.256} &                0.764 &                \textbf{0.858} &     0.546 &     \textbf{0.752} \\
 loudspeaker &  0.699 &  \textbf{0.807} &         0.360 &         \textbf{0.143} &                0.856 &                \textbf{0.912} &     0.801 &     \textbf{0.883} \\

       rifle &  0.466 &  \textbf{0.745} &         0.112 &         \textbf{0.012} &                0.789 &                \textbf{0.903} &     0.625 &     \textbf{0.851} \\
        sofa &  0.731 &  \textbf{0.809} &         0.171 &         \textbf{0.054} &                0.886 &                \textbf{0.927} &     0.831 &     \textbf{0.886} \\
       table &  0.569 &  \textbf{0.689} &         0.588 &         \textbf{0.058} &                0.873 &                \textbf{0.921} &     0.703 &     \textbf{0.805} \\
   telephone &  0.785 &  \textbf{0.861} &         0.103 &         \textbf{0.017} &                0.948 &                \textbf{0.971} &     0.866 &     \textbf{0.922} \\
      vessel &  0.592 &  \textbf{0.708} &         0.163 &         \textbf{0.053}&                0.818 &                 \textbf{0.868} &     0.730 &     \textbf{0.821} \\ \hline
        mean &  0.621 &  \textbf{0.749} &         0.265 &         \textbf{0.073} &                0.854 &                \textbf{0.906} &     0.743 &     \textbf{0.846} \\ \hline
\end{tabular}
\caption{Performance metrics for multi-view reconstruction using 5 random views of objects from categories seen at training time, where we see that 3D43D achieves consistently better reconstruction.}
\label{tab:seen_categories_5_view}
\end{table*}
\normalsize

\setlength{\tabcolsep}{1.5pt}
\setlength{\tabcolsep}{6pt}


\subsection{Generalization to unseen categories}
\label{sect:unknown_categories}

We now turn to our second experimental setup were we evaluate the ability of 3D43D to generalize to categories not seen during training. In order to do so, we restrict the training set to the top-3 most frequent categories in ShapeNet (\eg Car, Chair and Airplane) following \cite{unseenclasses}, and test on the remaining categories. Tab. \ref{tab:unknown_single_category_ccp} compares the performance of 3D43D with two strong baselines: OccNets \cite{occnet} and OccNets trained with a view-centric coordinate system (\cite{occnet}-v). We extend OccNets to use view-centric coordinates in order to validate observations in recent papers \cite{pixelsvoxelsviews, singleviewlearn} reporting that using a view-centric coordinate system improves generalization to unseen categories. We find empirically that this observation holds for models that do not aggregate information from multiple views. As discussed in Sec. \ref{sec:muli-view}, \cite{occnet}-v suffers from systematic drawbacks due to the use of global features, and this results in degraded performance. Additionally, using a view-centric coordinate system only partially tackles the generalization problem, and further improvements can be obtained from the geometry aware features, and the mean and variance aggregation used by 3D43D.

\begin{figure*}[!t]
\centering
\begin{tabular}{ccc}
\includegraphics[width=0.32\textwidth]{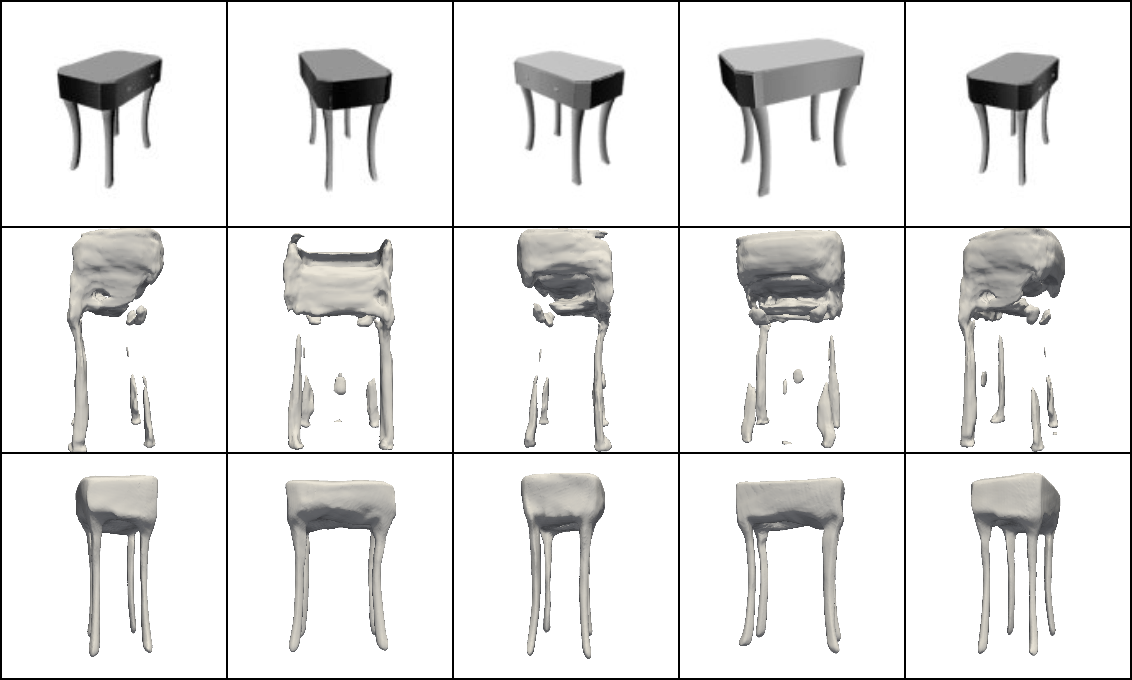} & \includegraphics[width=0.32\textwidth]{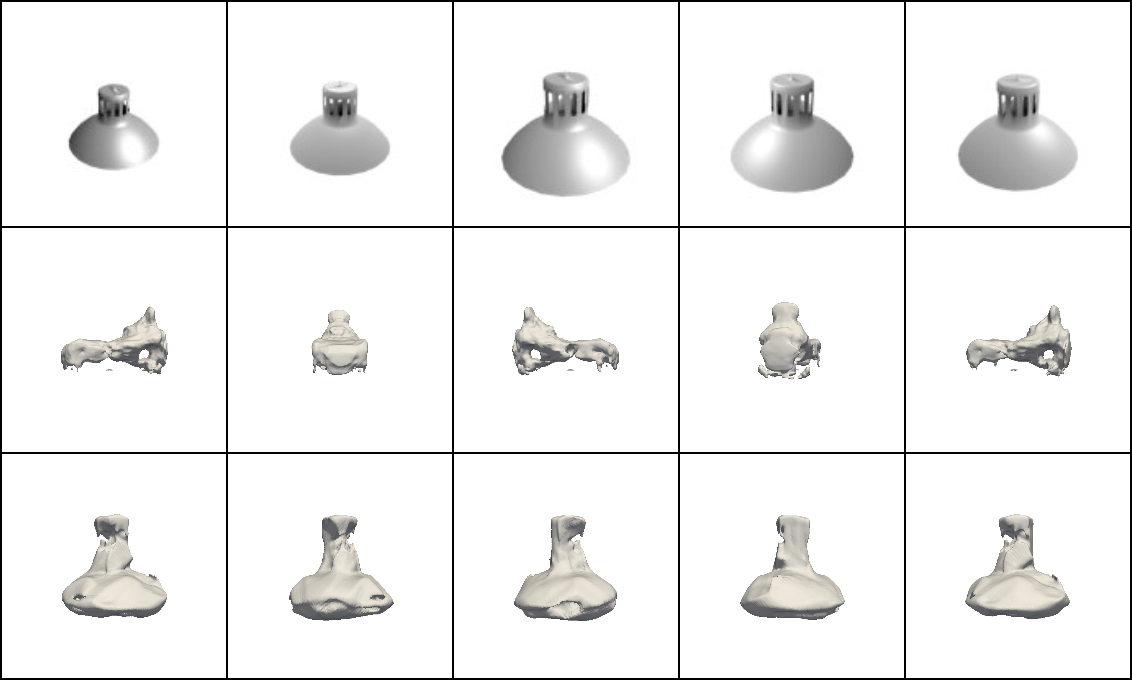} & \includegraphics[width=0.32\textwidth]{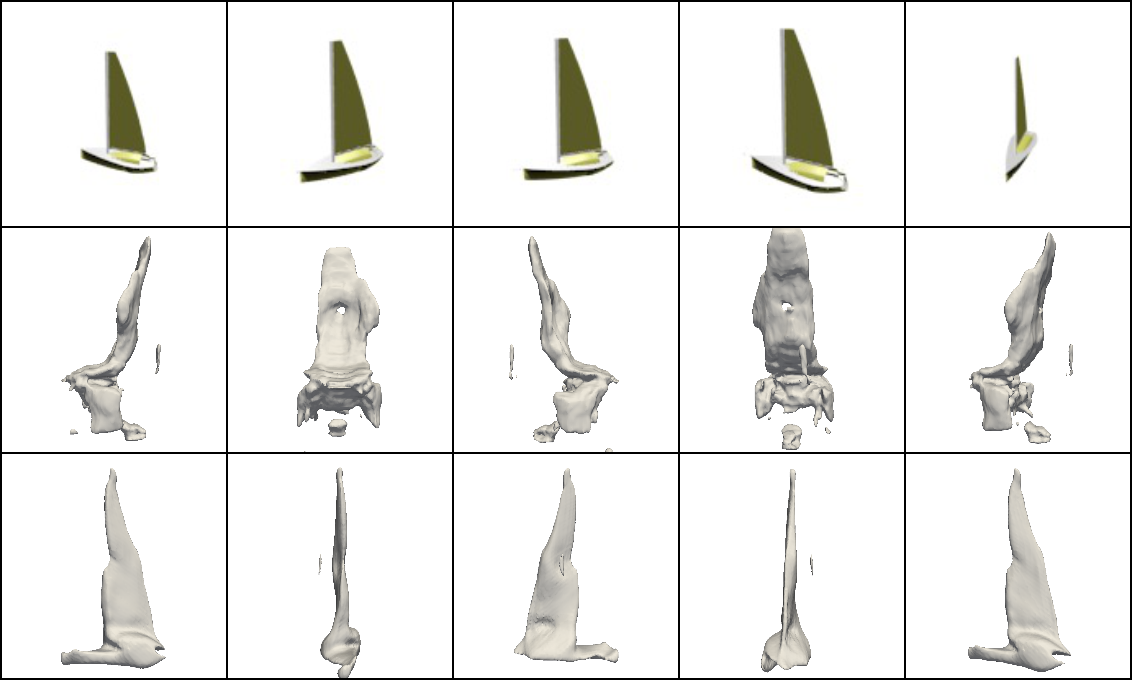} \\

\includegraphics[width=0.32\textwidth]{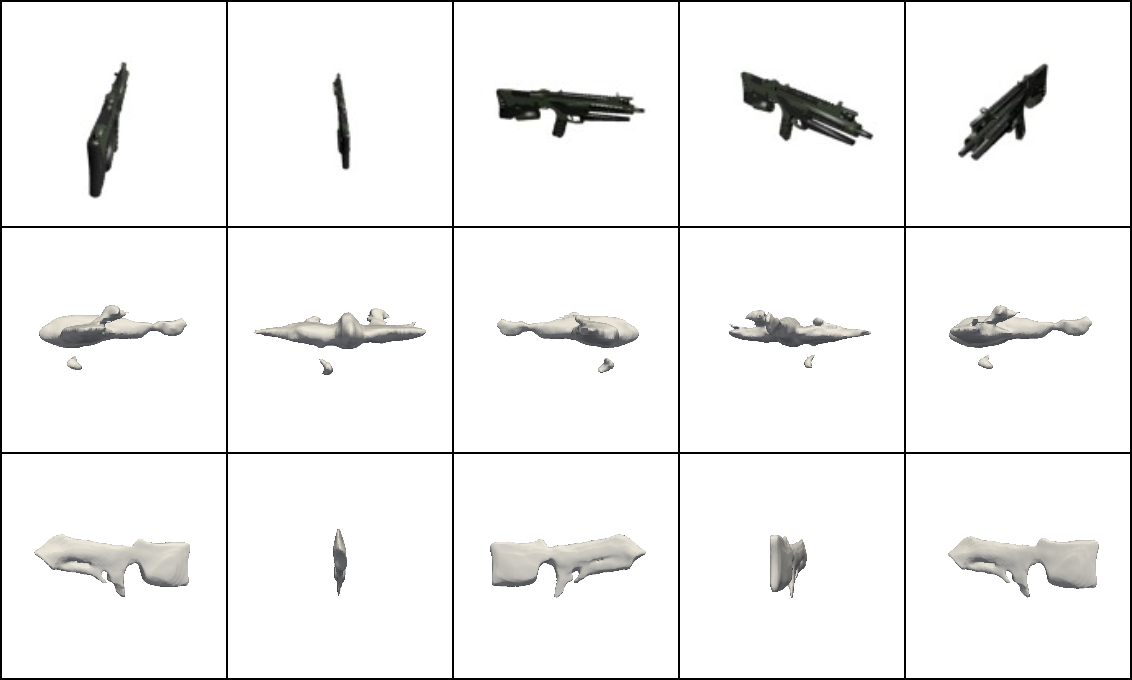} & \includegraphics[width=0.32\textwidth]{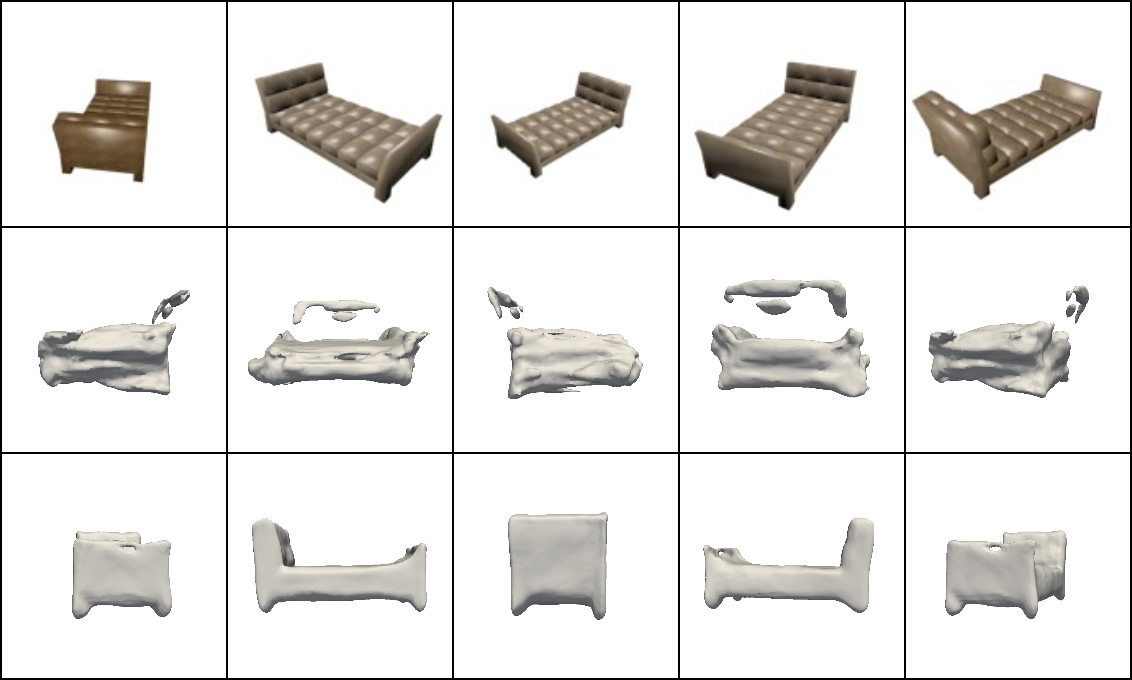} & \includegraphics[width=0.32\textwidth]{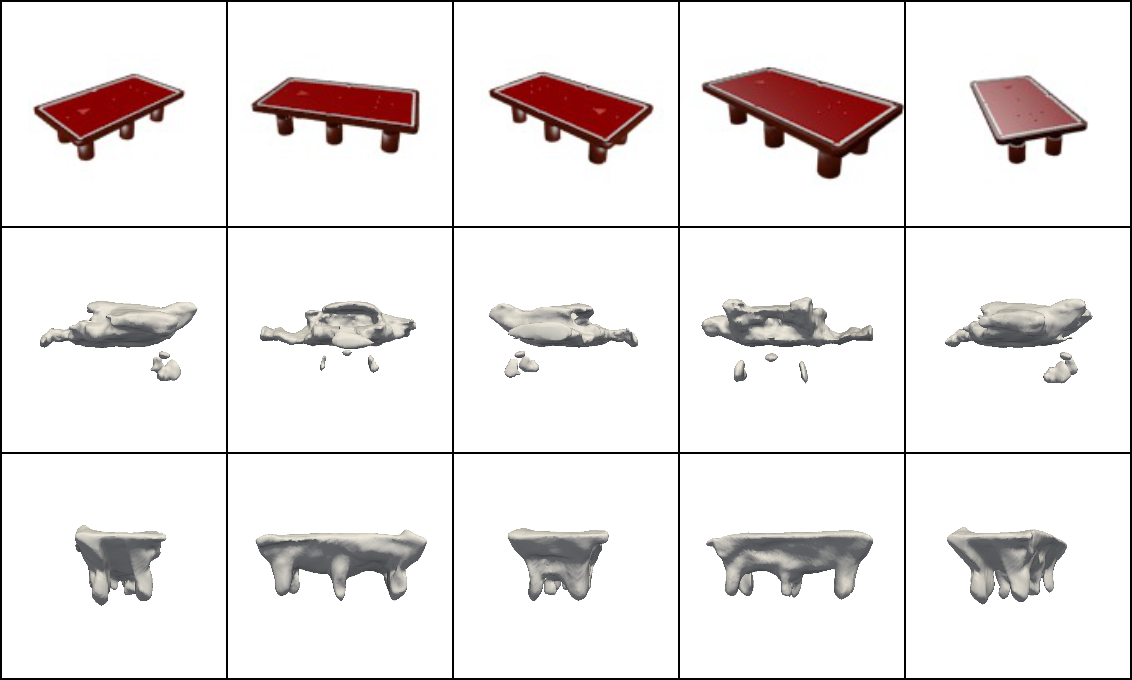} \\
\end{tabular}

\caption{Reconstruction of objects from unseen categories when training OccNets\cite{occnet} and 3D43D on \textit{Cars}, \textit{Chairs} and \textit{Airplanes}. For each object: (\textbf{Top row}) Input views. (\textbf{Middle row}) OccNets \cite{occnet} prediction (orbit of 5 views of the predicted mesh). (\textbf{Bottom row}) 3D43D prediction (orbit of 5 views of the predicted mesh). OccNets \cite{occnet} commonly map unseen categories to categories seen at training time. In comparison, 3D43D reconstructions are more accurate and less biased towards training categories. Note that input views and reconstructions are not shown from the same viewpoint.}
\label{fig:generalization}
\end{figure*}

We show sample reconstructions from this experiment in Fig. \ref{fig:generalization}. The visualizations reveal that OccNets tend to work in a categorization regime, often mapping unseen categories to their closest counterparts in the training set. This is clearly visible in Fig. \ref{fig:generalization}. This problem is not solved solely by using multiple views, which can be counterproductive by giving OccNets more confidence to reconstruct the wrong object.

\setlength{\tabcolsep}{1.5pt} 
\setlength{\tabcolsep}{6pt}

\begin{table*}
\scriptsize
\centering
\begin{tabular}{|l|ccc|ccc|ccc|ccc|}
\hline
& \multicolumn{3}{|c|}{\textbf{IoU}} & \multicolumn{3}{|c|}{$\downarrow$\textbf{Chamfer-}$L_1$} & \multicolumn{3}{|c|}{\textbf{Normal Consistency}} & 
\multicolumn{3}{|c|}{\textbf{F-score}}\\
\textbf{Unseen category}  & \cite{occnet} & \cite{occnet}-v & 3D43D & \cite{occnet} & \cite{occnet}-v & 3D43D & \cite{occnet} & \cite{occnet}-v & 3D43D & \cite{occnet} & \cite{occnet}-v & 3D43D \\
\hline
\textit{(1 view)} & & & & & & & & & & & & \\
bench   &   0.251   &   0.291  &   \textbf{0.302} &   0.752  &   \textbf{0.323} &   0.357  &   0.714  &   \textbf{0.733} &    0.706 &   0.374  &   0.426  &   \textbf{0.447} \\
cabinet &   0.282  &   0.404  &   \textbf{0.502} &   1.102  &   0.621  &   \textbf{0.529} &   0.662  &   0.739 &   \textbf{0.759}  &   0.418  &   0.551  &   \textbf{0.647} \\
display &   0.117  &   0.162  &   \textbf{0.243} &   3.213  &   1.836  &   \textbf{1.389} &   0.546  &   0.612  &   \textbf{0.638} &   0.197  &   0.260  &   \textbf{0.364} \\
lamp    &   0.100  &   0.150  &   \textbf{0.223} &   3.482  &   2.276  &   \textbf{1.997} &   0.582  &   \textbf{0.625}  &   0.618 &   0.166  &   0.241  &   \textbf{0.340}\\
loudspeaker &   0.311  &   0.405  &   \textbf{0.507} &   1.649  &   0.860  &   \textbf{0.744} &   0.655  &   0.731  &   \textbf{0.749} &   0.452  &   0.552  &   \textbf{0.649} \\
rifle   &   0.155  &   0.150  &   \textbf{0.236} &   2.465  &   2.206  &   \textbf{0.707} &   0.539  &   0.527  &   \textbf{0.588} &   0.255  &   0.252  &   \textbf{0.3737} \\
sofa    &   0.493  &   0.552  &   \textbf{0.559} &   0.915  &   \textbf{0.399}  &   0.421 &   0.761  &   \textbf{0.799}  &   0.784 &   0.625  &   0.688  &   \textbf{0.699} \\
table   &   0.172  &   0.214  &   \textbf{0.313} &   1.304  &   0.861  &   \textbf{0.583} &   0.686  &   0.722  &   \textbf{0.731} &   0.275  &   0.331  &   \textbf{0.461} \\
telephone &  0.052 &   0.155  &   \textbf{0.271} &   1.673  &   1.062  &   \textbf{0.996} &   0.654  &   0.682  &   \textbf{0.700}  &   0.096  &   0.256  &   \textbf{0.403} \\
vessel  &   0.324  &   0.378  &   \textbf{0.401} &   0.849  &   0.592  &   \textbf{0.521} &   0.648  &   \textbf{0.691} &   0.690  &   0.463  &   0.525  &    \textbf{0.553} \\ \hline
mean    &   0.226  &   0.286  &   \textbf{0.356} &   1.740  &   1.104  &   \textbf{0.824} &   0.645  &   0.686  &   \textbf{0.696} &   0.332  &   0.408  &    \textbf{0.494} \\

\hline
\textit{(5 views)} & & & & & & & & & & & & \\

bench   &   0.288  &   0.147  &   \textbf{0.463} &   0.508  &   1.960  &   \textbf{0.113} &   0.729  &   0.625  &   \textbf{0.800} &   0.421  &   0.242  &   \textbf{0.617} \\
cabinet &   0.295  &   0.312  &   \textbf{0.629} &   0.917  &   1.273  &   \textbf{0.250} &   0.674  &   0.655  &   \textbf{0.844} &   0.430  &   0.458  &   \textbf{0.756} \\
display &   0.120  &   0.127  &   \textbf{0.409} &   2.868  &   3.179  &   \textbf{0.428} &   0.560  &   0.534  &   \textbf{0.770} &   0.200  &   0.213  &   \textbf{0.558} \\
lamp    &   0.100  &   0.138  &   \textbf{0.369} &   3.365  &   2.653  &   \textbf{2.057}  &   0.586  &   0.623  &   \textbf{0.738} &   0.167  &   0.224  &   \textbf{0.513} \\
loudspeaker &   0.315  &   0.333  &   \textbf{0.627} &   1.460  &   1.344  &   \textbf{0.392} &   0.660  &   0.677  &   \textbf{0.829} &   0.457  &   0.480  &   \textbf{0.753} \\
rifle   &   0.180  &   0.095  &   \textbf{0.498} &   1.866  &   2.610  &   \textbf{0.115} &   0.567  &   0.444  &   \textbf{0.760} &   0.290  &   0.169  &   \textbf{0.655} \\
sofa    &   0.525  &   0.356  &   \textbf{0.679} &   0.732  &   1.445  &   \textbf{0.147} &   0.776  &   0.663  &   \textbf{0.858} &   0.656  &   0.508  &    \textbf{0.795} \\
table   &   0.186  &   0.177  &   \textbf{0.455} &   1.122  &   1.771  &   \textbf{0.255} &   0.694  &   0.700  &   \textbf{0.827} &   0.295  &   0.285  &    \textbf{0.609} \\
telephone   &   0.036  &   0.131  &   \textbf{0.549} &   1.588  &   1.457  &   \textbf{0.184} &   0.689  &   0.592  &   \textbf{0.861} &   0.066  &   0.226  &   \textbf{0.691} \\
vessel  &   0.347  &   0.256  &   \textbf{0.521} &   0.683  &   1.524  &   \textbf{0.145} &   0.661  &   0.603  &   \textbf{0.776} &   0.489 &   0.390  &   \textbf{0.669} \\ \hline
mean    &   0.239  &   0.207  &   \textbf{0.520} &   1.511  &   1.922  &   \textbf{0.409} &   0.660  &   0.612  &   \textbf{0.806} &   0.347  &   0.319  &   \textbf{0.662} \\ \hline

\end{tabular}
\caption{Performance metrics for single and multi-view reconstruction when generalizing to unseen object categories after training only on the \textbf{car}, \textbf{chair} and \textbf{plane} categories.}
\label{tab:unknown_single_category_ccp}
\end{table*}

\normalsize

\subsection{Ablation}


We perform an ablation study to show how the main design choices of our approach affect performance (ie. a point representation that is aware of camera position and a variance cost to aggregate information across views). We take as our baseline a model conceptually similar to DISN \cite{disn} but with a view-centric coordinate frame. We have already shown that spatial feature maps provide substantial improvements over 1D features (\eg improvements over OccNet \cite{occnet} shown in Tab. (\ref{tab:all_category})(\ref{tab:seen_categories_5_view})) for seen categories. Note that DISN \cite{disn} also reports similar results. However, in this paper we focus on analyzing the generalization of the model to categories unseen during training and show in our ablation that spatial feature maps are not the only critical design choice and our novel contributions improve reconstruction accuracy for unseen categories.

For all our model ablations our encoder (ie. a UNet with a ResNet50 encoder) outputs spatial feature maps for each view that are sampled at locations corresponding to a particular point $\textbf{p} \in \mathbb{R}^3$ for which occupancy is predicted. Our ablation is divided in three models:

\begin{itemize}
    \item Point model (P): Here we take the sampled features across views (\eg. the $\textbf{c}_i$) and concatenate them with $\textbf{p}$ before feeding through our MLP $g_\theta$, so that $\textbf{g}_i = g_\theta(\textbf{c}_i, \textbf{p})$. We take these feature representations $\textbf{g}_i$ across views and aggregate them using average pooling, where the resulting vector is used as input to $f_\theta$.
    
    \item Point+Camera model (P+C): In this version we concatenate also the camera location $\textbf{t}_i$ before processing the vector with $g_\theta$, so that $\textbf{g}_i = g_\theta(\textbf{c}_i, \textbf{p}, \textbf{t}_i)$. We then average pool the resulting features and use them as input to $f_\theta$.
    
    \item Point+Camera+Variance model (P+C+V): In this model we take the same encoding as in P+C ($\textbf{g}_i = g_\theta(\textbf{c}_i, \textbf{p}, \textbf{t}_i)$). However, we now compute the mean and variance of $\textbf{g}_i$ and use them as input and conditioning, respectively, for $f_\theta$. This is our full 3D43D model.
\end{itemize}

These models are trained on 3 ShapeNet categories: plane, chair and car. We then report results on the test set of 10 unseen categories.  We train and evaluate our model with $4$ views and report the average IoU across the unseen classes in Tab. \ref{tab:ablation}, where we show that our novel contributions contribute to improve the reconstruction accuracy.

\begin{table}[!h]
    \small
    \centering
    \begin{tabular}{|c|c|c|c|}
        \hline
         & P & P+C & P+C+V (3D43D) \\ \hline
         IoU &	0.453 &	0.476	& \textbf{0.491} \\
         \hline
    \end{tabular}
    \caption{Results of our ablation experiments.}
    \label{tab:ablation}
\end{table}

\section{Conclusions}

In this paper, we studied factors that impact the generalization of learning-based 3D reconstruction models to unseen categories during training. We argued that for a 3D reconstruction approach to generalize successfully to unseen classes all these factors need to be addressed together. We empirically showed that by taking this into when designing our model, we obtain large improvements over state-of-the-art methods when reconstructing objects of on unseen categories. These improvements in generalization are a step forward for learned 3D reconstruction methods, which we hope will also enable recent Neural Rendering approaches \cite{nerf,srn,enr} to go beyond the constrained scenario of training-category specific models. Finally, larger datasets will lead to more informative priors. We believe that having a clear understanding of these factors and their compound effects will enrich this promising avenue of research.

{\small
\bibliographystyle{ieee_fullname}
\bibliography{egbib.bib}
}
\end{document}